\documentclass[journal]{IEEEtran}

\usepackage{amsmath}
\usepackage{amsfonts}
\usepackage{amssymb}
\usepackage{bm}

\def\*#1{\mathbf{#1}}
\def\+#1{\mathcal{#1}}
\def\-#1{\mathbb{#1}}
\def\~#1{\mathrm{#1}}

\newcommand{\method}{EPC-RawDC}

\usepackage{calc}
\usepackage{cite}
\usepackage[ruled]{algorithm2e}

\usepackage{graphicx}
\usepackage{textcomp}
\usepackage[table]{xcolor}
\usepackage{multirow}
\usepackage{microtype} 

\usepackage[single]{acro}

\makeatletter
\let\NAT@parse\undefined
\makeatother

\usepackage[
    colorlinks=true,
    linkcolor=blue,
    citecolor=blue,
    urlcolor=blue
]{hyperref}

\usepackage{pgfplots}
\usepackage{pgf}
\usepackage{tikz}
\usetikzlibrary{matrix}
\usetikzlibrary{positioning, arrows.meta, calc, spy}

\usetikzlibrary{fit}					  
\usetikzlibrary{backgrounds}	         
\usetikzlibrary{shapes.arrows}
\usetikzlibrary{shapes.geometric}
\usetikzlibrary{shapes.multipart}
\usetikzlibrary{matrix, arrows, positioning}
\tikzstyle{rec}=[draw,rectangle, minimum height=2cm]
\tikzset{>=stealth', punkt/.style={
		fill=gray!40, 
		draw=black, very thick, text width=4.7em, minimum height=3em, text centered}}
\tikzset{>=stealth', Denoi/.style={rectangle, rounded    corners=2mm, fill=blue!20, 
		draw=black, very thick, text width=5.0em, minimum height=3.5em, text centered}}
\tikzset{>=stealth', CG/.style={rectangle, 
		fill=green!20, draw=black, very thick, text width=9.5em, minimum height=3.5em, text centered}}

\tikzstyle{background} = [rectangle, fill=green!20, inner sep=0.1cm, rounded corners=4mm, 
minimum height=6em]
\tikzstyle{sum}   = [draw, fill=gray!40, circle, node distance=1cm]
\tikzstyle{dot}   = [circle, fill=black, inner sep=0pt, minimum size=5pt, node contents={}]
\tikzstyle{fig_n} = [node distance=30pt, inner sep=0cm]

\tikzset{
    physop/.style={
        rectangle,
        rounded corners=2mm,
        fill=orange!20,
        draw=black,
        very thick,
        text width=4em,
        minimum height=3.2em,
        align=center
    },
    eqloss/.style={
        rectangle,
        rounded corners=2mm,
        fill=blue!15,
        draw=blue!60!black,
        very thick,
        text width=12em,
        minimum height=4.0em,
        align=center
    },
    rawloss/.style={
        rectangle,
        rounded corners=2mm,
        fill=green!18,
        draw=green!50!black,
        very thick,
        text width=11em,
        minimum height=4.0em,
        align=center
    },
    latent/.style={
        rectangle,
        rounded corners=2mm,
        fill=cyan!25,
        draw=black,
        thick,
        text width=5.0em,
        minimum height=3em,
        align=center
    }
}

\def\BibTeX{{\rm B\kern-.05em{\sc i\kern-.025em b}\kern-.08em
    T\kern-.1667em\lower.7ex\hbox{E}\kern-.125emX}}
\begin{document}
	
\title{EPC-RawDC: Equivariant Physics Constrained 3D Diffusion with Raw Data Consistency for CBCT to CT Image Synthesis}

\author{Alzahra Altalib, Chunhui Li, Alessandro Perelli 
\thanks{This work involved human subjects in its research. All ethical and experimental procedures for the JUST dataset were approved with protocol Mar2026/191-38 at King Abdullah University Hospital (Jordan).}
\thanks{A. Altalib was supported by the Jordan University of Science and Technology PhD scholarship and A. Perelli by the Royal Academy of Engineering / Leverhulme Trust Research Fellowship LTRF-2324-20-160.} 
\thanks{A. Altalib and C. Li are with the Faculty of Science, Engineering and Business, University of Dundee, DD1 4HN, UK.}
\thanks{A. Altalib is with the Faculty of Applied Medical Sciences, Jordan University of Science and Technology, Irbid, 21410, Jordan} 
\thanks{A. Perelli is with the School of Cardiovascular and Metabolic Health, University of Glasgow, G12 8TA (UK).} 
\thanks{\hspace*{-1.1em}Corresponding author: Alzahra Altalib} 
\thanks{\hspace*{-1.1em}Email: \mbox{\texttt{2600129@dundee.ac.uk}}}
}

\maketitle

\begin{abstract}
Cone beam computed tomography (CBCT) is widely used for image guided and adaptive radiotherapy, but scatter, noise, reconstruction artifacts, and inaccurate Hounsfield units limit its quantitative use. Recent diffusion approaches for CBCT to CT synthesis remain largely image driven, with limited use of the underlying acquisition physics. We propose EPC-RawDC, a conditional 3D latent diffusion framework that combines volumetric generative modeling with two complementary constraints derived from the same scanner specific CBCT forward model. 
First, geometry equivariant projection constraint exploits the rotational equivariance of the CBCT forward operator by supervising the synthesized CT through correspondingly transformed reference projections. Second, raw data consistency uses noisy projection measurements of the individual acquisition as scan specific supervision during training. The framework was evaluated on two independent clinical cohorts with distinct scanner geometries and projection sources. For JUST, the physics constraints used measured clinical raw CBCT projections acquired with an Elekta system, whereas for SynthRAD2025 the projection data were simulated from the reference CT using a Varian geometry with Poisson photon statistics. EPC-RawDC achieved an SSIM of 0.97 in both cohorts and improved PSNR over DDPM by 2.66 dB on SynthRAD2025 and 2.10 dB on JUST, while reducing mean MAE by 20.3\% and 19.0\%, respectively. On SynthRAD2025, projection consistency and equivariant errors were reduced by 55.0\% and 41.4\% relative to DDPM. These results demonstrate that combining a volumetric diffusion prior with scanner geometry and scan specific projection measurements improves quantitative CBCT to CT synthesis across both simulated and measured projection domains. Code available: \href{GitHub}{https://github.com/ALZAHRAALTALIB/EPC-RawDC}.
\end{abstract}


\section{Introduction} \label{sec:introduction}

Cone beam computed tomography (CBCT) is widely used for image guidance and adaptive radiotherapy because it provides volumetric information directly at the treatment position. Its quantitative use, however, remains limited by scatter, beam hardening, photon noise, truncation, detector effects, and reconstruction artifacts. These degradations reduce soft tissue contrast and disturb the relationship between reconstructed intensity and tissue attenuation, resulting in inaccurate Hounsfield units (HU) and limiting applications such as dose calculation and online treatment adaptation. 
Synthetic CT (sCT) generation from CBCT has therefore become an important approach for obtaining CT like quantitative information from daily treatment imaging. The SynthRAD2025 challenge demonstrated substantial progress in this area using large multicenter datasets, while also showing that image similarity alone does not necessarily predict dosimetric accuracy \cite{huijben2024synthrad}. This motivates methods that preserve image quality while incorporating stronger knowledge of the underlying acquisition process.

Early learning based CBCT to CT methods relied primarily on convolutional networks and generative adversarial models. Cycle consistent architectures enabled unpaired image translation and substantially improved HU accuracy over uncorrected CBCT \cite{liang2019cyclegan}, while more recent approaches have incorporated transformer architectures, anatomical constraints, and uncertainty estimation \cite{rusanov2024transformer}. Diffusion models have subsequently emerged as a powerful alternative because of their stable denoising objective and ability to model complex conditional image distributions. Peng \emph{et al.} demonstrated conditional denoising diffusion for CBCT based CT generation with improved image quality and HU fidelity \cite{peng2024cbctddpm}. More recent studies have explored anatomical guidance, frequency information, patient adaptation, latent representations, and external clinical validation \cite{altalib2026review}. 3D latent diffusion has also been investigated for volumetric CBCT to CT synthesis \cite{alshalabi2026ldm}, while multicenter conditional diffusion studies have shown the feasibility of validation across independent clinical populations \cite{xie2026multicenter}. Consequently, neither latent diffusion nor multicenter evaluation alone constitutes the principal novelty of the present work.

Despite these advances, most CBCT to CT synthesis methods remain predominantly image driven. A generated CT can be anatomically plausible and visually realistic without necessarily satisfying the physical relationships imposed by the CBCT acquisition system or remaining compatible with the projection measurements from which the input CBCT was reconstructed. Latent diffusion provides an efficient mechanism for modeling complete three dimensional volumes \cite{rombach2022ldm}, but the learned generative prior alone does not guarantee consistency with scanner geometry or patient specific measurements. This limitation is particularly relevant to CBCT, where the acquisition trajectory and forward projection model provide strong physical structure that is known independently of the training data.

Equivariance provides one mechanism for incorporating this information. Equivariant imaging demonstrated that symmetries of an imaging operator can constrain learned inverse problems beyond conventional image domain objectives \cite{chen2021equivariant}. In CBCT to CT synthesis, EqDiff-CT introduced rotational equivariance at the network representation level \cite{altalib2026eqdiff}. 
Our previous EPC-3D-Diff framework instead derived a projection domain constraint from the rotational equivariance of the CBCT acquisition operator \cite{altalib2026epc3d}. An in plane rotation of the three dimensional object about the scanner axis produces a corresponding angular displacement of its projections. EPC-3D-Diff used this scanner derived transformation law to construct geometry transformed projection supervision within a conditional three dimensional latent diffusion model, rather than generic image domain rotation augmentation. 

The geometry derived term in EPC-3D-Diff uses a noise free reference projection transformed according to the deterministic symmetry of the scanner operator; it does not incorporate the noisy projection realization measured for an individual scan. The present work addresses this limitation by introducing raw data consistency (RawDC). Both terms use the same scanner specific forward model, but RawDC adds patient and scan specific measurement supervision that is absent from the geometry transformed reference projection term. The acquired photon counts originate from Poisson statistics, while the RawDC objective uses the corresponding post-log quadratic Gaussian approximation developed in Section~\ref{subsec:measurement_model}. Importantly, the implemented forward model preserves scanner specificity by evaluating projections at the native detector resolution of the corresponding acquisition system.

\subsection{Positioning Against Existing Methods} \label{subsec:positioning}

Existing CBCT to CT synthesis methods based on GANs, transformers, and conditional diffusion primarily learn image domain mappings \cite{liang2019cyclegan,rusanov2024transformer,peng2024cbctddpm,altalib2026review} and volumetric latent diffusion improves the efficiency of three dimensional modeling \cite{alshalabi2026ldm}. Projection consistency has also been incorporated into tomographic reconstruction methods \cite{zhou2021projection,chen2025cddm}, but these approaches reconstruct images directly from projection data rather than perform conditional CBCT to CT synthesis. 
EPC-RawDC combines a volumetric latent diffusion prior, geometry transformed projection supervision derived from scanner equivariance, and scan specific raw projection consistency within a single synthesis framework.  

Relative to EqDiff and EPC-3D-Diff~\cite{altalib2026eqdiff,altalib2026epc3d}, the present work introduces scan specific raw data consistency that directly constrains the synthesized CT using the projection measurements of the corresponding acquisition. 

\subsection{Main Contributions} \label{subsec:contributions}

The principal novelty of EPC-RawDC is the integration of an equivariance derived projection constraint with scan specific raw projection consistency within a common acquisition CBCT forward model for the conditional 3D latent diffusion. 

The contributions are threefold. First, we formulate conditional 3D latent diffusion for volumetric CBCT to CT synthesis. 
Second, we impose geometry transformed projection supervision using the physical relationship between object rotation and projection angle shift induced by the scanner operator. Third, RawDC constrains the predicted CT during training using the projection measurements of the corresponding acquisition. Both physics terms use the same acquisition faithful forward operator while encoding complementary reference geometry and scan specific measurement information. The framework is evaluated on independent datasets with distinct scanner geometries using controlled ablations of volumetric diffusion, geometry derived projection supervision, and raw data consistency. 

Together, these contributions establish EPC-RawDC as a measurement grounded generative framework in which the sCT is constrained jointly by the learned CT distribution, the known scanner geometry, and the projection measurements of the individual acquisition.

\section{Physics Based Formulation}
\label{sec:physics_formulation}

\subsection{Statistical CT and CBCT Measurement Model}\label{subsec:measurement_model}

Following the notation of EPC-3D-Diff, let $\*x_{r}$ denote the attenuation volume and $\*A_{r}$ the corresponding X-ray forward operator, with $r\in\{0,c\}$ indicating CT and CBCT, respectively. Under the
Beer--Lambert model, the detected photon counts are modeled as
\begin{equation}\label{eq:poisson_model}
\*p_{r}\mid\*x_{r} \sim \operatorname{Poisson} \left(\*I_{r}\odot\exp(-\*A_{r}\*x_{r})\right),
\end{equation}
where $\*I_{r}$ denotes the incident photon fluence and $\odot$ denotes elementwise multiplication. After logarithmic normalization, 
\begin{equation}\label{eq:log_measurement}
\*y_r = -\log\left(\*p_{r}\oslash\*I_{r}\right),
\end{equation}
where $\oslash$ denotes elementwise division. For sufficiently high photon counts, applying the Gaussian approximation to the Poisson distribution together with a first order approximation of the logarithmic transform gives 
\begin{equation}\label{eq:postlog_gaussian}
\*y_r \approx \*A_r\*x_r + \*n_r, \qquad \*n_r \sim \mathcal{N}\left(\*0, \boldsymbol{\Sigma}_{r}\right),
\end{equation}
with approximately signal dependent variance $\sigma_{r,j}^{2}\simeq1/\bar{p}_{r,j}$. Defining $\*W_r=\boldsymbol{\Sigma}_{r}^{-1}$, the corresponding Gaussian negative log likelihood reduces, up to additive and multiplicative constants, to the weighted least squares data fidelity 
\begin{equation}\label{eq:wls}
\mathcal{L}_{\mathrm{WLS}} = \left\| \*W_r^{1/2} \left(\*y_{r} - \*A_r\*x_{r}\right)\right\|_{2}^{2}.
\end{equation}

Thus, although the underlying photon statistics are Poisson, the post-log measurements admit a convenient heteroscedastic Gaussian approximation, providing the statistical basis for quadratic projection domain consistency terms \cite{nuyts2013physics,pwlsCT}. When uniform weighting is adopted $\*W_{r}=\*I$, Eq.~\eqref{eq:wls} reduces to the standard least squares objective.

\subsection{CBCT Geometry Induced Projection Equivariance} \label{subsec:cbct_equivariance}

Equivariance in the proposed framework is not introduced as a generic image augmentation prior. Instead, it follows directly from the rotational symmetry of the CBCT acquisition geometry. In a circular CBCT scan, the X-ray source and detector rotate around a fixed scanner axis while maintaining their relative source, isocenter, and detector geometry \cite{chen2021equivariant}. Consequently, rotating the object around the axis produces an equivalent angular displacement of its projection data.

Let $\mathbf{R}_{\phi}$ denote a three dimensional rotation of an attenuation volume by angle $\phi$ about the CBCT rotation axis, and let $\mathbf{A}_{c}$ denote the CBCT forward projector at the native detector resolution. For a projection acquired at gantry angle $\eta$, the continuous circular acquisition geometry satisfies
\begin{equation}\label{eq:cbct_rotation_property}
\left[ \mathbf{A}_{c} \mathbf{R}_{\phi}\mathbf{x} \right](\eta,u,v) = \left[\mathbf{A}_{c}\mathbf{x}\right](\eta-\phi,u,v),
\end{equation} 
where $(u,v)$ are detector coordinates. Defining the angular translation operator $\left[\mathbf{T}_{\phi}\mathbf{y}\right](\eta,u,v)
:=\mathbf{y}(\eta-\phi,u,v)$, Eq.~\eqref{eq:cbct_rotation_property} gives the scanner geometry induced relation
\begin{equation}\label{eq:full_equivariance}
\mathbf{A}_{c}\mathbf{R}_{\phi} = \mathbf{T}_{\phi}\mathbf{A}_{c}.
\end{equation}
Equation~\eqref{eq:full_equivariance} is an intertwining relation between rotations in the image domain and angular translations in the projection domain. It characterizes equivariance of the CBCT acquisition operator, is determined by the circular scanner geometry rather than learned from the training data, and forms the physical basis of the geometry derived projection constraint introduced in Section~\ref{subsec:eq_loss}.  

In the ideal continuous circular geometry, Eq.~\eqref{eq:full_equivariance} is exact. In the discrete implementation it is satisfied approximately because the projection angles, detector grid, volume and rotational interpolation are discretized. Accordingly, $\mathbf{T}_{\phi}$ is implemented as an integer angular shift when $\phi$ coincides with the acquisition sampling and by fractional angular interpolation otherwise. 

Importantly, Eq.~\eqref{eq:full_equivariance} describes the acquisition operator and does not by itself imply equivariance of the learned CBCT to CT mapping. In the proposed training objective, this operator relation is used to construct geometry transformed reference projection supervision, while RawDC introduces scan specific measurement supervision. The two terms therefore use the same scanner model but constrain the prediction with different projection domain targets.

\section{Proposed EPC-RawDC Framework} \label{sec:proposed_framework}

\begin{figure*}[!t]
\centering

\resizebox{\textwidth}{!}{
\begin{tikzpicture}[node distance=2.2cm, scale=1]


\node[
    fig_n,
    inner sep=0pt,
    label=above:{Reference CT}
] (CT) at (-2.6,1.3)
{
    \includegraphics[width=.12\textwidth]{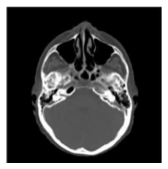}
};

\node[
    punkt,
    fill=yellow!30,
    trapezium,
    shape border rotate=270,
    align=center,
    text width=2.1em,
    minimum height=1.2em,
    text height=3em,
    text depth=20pt
] (ECT) at (-0.5,1.3)
{
    $\mathcal{E}$
};

\draw[->,very thick]
(CT.east) -- node[above] {$\*x_0$} (ECT.west);

\node[latent] (z0) at (2.4,1.3) { $\*z_0$ };

\draw[->,very thick]
(ECT.east) -- node[above] {Encoder} (z0.west);

\node[latent] (ztm1) at (4.7,1.3) { $\*z_{T-1}$ };

\node[latent] (zT) at (7.0,1.3) { $\*z_{T}$ };

\draw[->,dashed,very thick] (z0.east) -- (ztm1.west);

\draw[->,very thick] (ztm1.east) -- (zT.west);

\node[
    fig_n, inner sep=0pt,
    label=above:{\shortstack{Gaussian\\latent noise}}
] (noise) at (9.4,1.3)
{ \includegraphics[width=.11\textwidth]{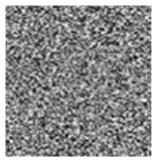} };

\draw[->,very thick] (zT.east) -- (noise.west);


\node[Denoi] (epsT) at (7.0,-1.8)
{
    $\boldsymbol{\epsilon}_{\theta}
    (\mathbf{z}_{T}, \mathbf{z}_{c})$
};

\node[Denoi] (epsTm1) at (4.7,-1.8)
{
    $\bm\epsilon_{\theta} (\*z_{T-1}, \*z_{c})$
};

\node[Denoi] (eps1) at (2.4,-1.8)
{
    $\bm\epsilon_{\theta} (\*z_{1}, \*z_{c})$
};

\draw[->,very thick] (noise.south) |- (epsT.east);

\draw[->,very thick] (epsT.west) -- (epsTm1.east);

\draw[->,dashed,very thick] (epsTm1.west) -- (eps1.east);

\node[
    punkt,
    fill=yellow!30,
    trapezium,
    shape border rotate=90,
    align=center,
    text width=2.1em,
    minimum height=1.2em,
    text height=3em,
    text depth=20pt
] (DCT) at (-0.5,-1.8)
{
    $\mathcal{D}$
};

\draw[->,very thick]
(eps1.west) --
node[above] {$\hat{\*z}_0$}
node[below] {Decoder}
(DCT.east);

\node[
    fig_n, inner sep=0pt, label=above:{Predicted sCT}
] (sCT) at (-2.6,-1.8)
{
    \includegraphics[width=.12\textwidth]{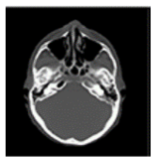}
};

\draw[->,very thick] (DCT.west) -- node[above] {$\hat{\mathbf{x}}_0$} (sCT.east);


\node[
    fig_n, inner sep=0pt,
    label={[yshift=-4pt]below:{\shortstack{Noisy CBCT\\projection data $\*y_c$}}}
] (yc) at (4.4,-5.3)
{ \includegraphics[width=.17\textwidth]{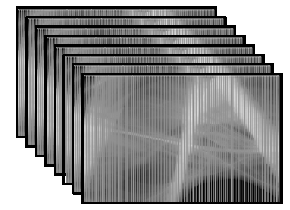} };

\node[
    fig_n, inner sep=0pt, label=below:{CBCT image}
] (CBCT) at (9.0,-5.3)
{ \includegraphics[width=.12\textwidth]{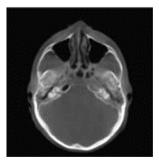} };

\draw[->,very thick] (yc.east) -- node[above] {\shortstack{CBCT\\reconstruction}} (CBCT.west);

\node[
    punkt,
    fill=yellow!30,
    trapezium,
    shape border rotate=0,
    align=center,
    text width=2.1em,
    minimum height=2.2em
] (ECBCT) at (9.0,-3.3)
{
    $\mathcal{E}$
};

\draw[->,very thick] (CBCT.north) -- node[right] {$\*x_c$}
(ECBCT.south);

\coordinate (zcbus) at (7.0,-3.3);

\draw[-,very thick] (ECBCT.west) -- node[above] {$\*z_c$}
(zcbus);

\draw[->,very thick] (zcbus) |- (epsT.south);

\coordinate (zcbus1) at (4.7,-3.3);

\draw[-,very thick] (ECBCT.west) -- (zcbus1);

\draw[->,very thick] (zcbus1) |- (epsTm1.south);

\coordinate (zcbus2) at (2.4,-3.3);

\draw[-,very thick] (ECBCT.west) -- (zcbus2);

\draw[->,very thick] (zcbus2) |- (eps1.south);


\node[physop] (GcRef) at (-6.2,1.3)
{
    $\*A_c$
};

\draw[->,very thick]
(CT.west) -- node[above] {$\mathbf{x}_0$} (GcRef.east);

\node[
    fig_n, inner sep=0pt, label={[yshift=2pt]above:{
        $\*T_{\phi}(\bar{\*y}_{0\rightarrow c})$
    }}
] (Ty) at (-10.8,1.3)
{
    \includegraphics[width=.19\textwidth]{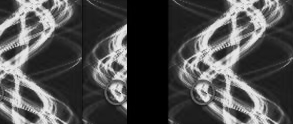}
};

\draw[->,very thick]
(GcRef.west) -- node[above] {$\mathbf{T}_{\phi}$} (Ty.east);

\node[
    fig_n,
    inner sep=0pt,
    label={[yshift=-4pt]below:{
        $\mathbf{R}_{\phi}
        (\hat{\mathbf{x}}_0)$
    }}
] (sCTrot) at (-5.7,-1.8)
{
    \rotatebox{45}{
        \includegraphics[width=.12\textwidth]{images/sCT.png}
    }
};

\draw[->,very thick]
(sCT.west) -- node[above] {$\*R_{\phi}$} (sCTrot.east);

\node[physop] (GcEq) at (-9.0,-2.7)
{
    $\*A_c$
};

\draw[->,very thick] (sCTrot.west) ++(1,-.9) -- (GcEq.east);

\node[eqloss] (Leq) at (-10.8,-.6)
{
    $\displaystyle
    \mathcal{L}_{\mathrm{eq}} = \left\| \*A_c
    \mathbf{R}_{\phi} (\hat{\mathbf{x}}_0) - \mathbf{T}_{\phi}
    (\bar{\mathbf{y}}_{0\rightarrow c}) \right\|_2^2$
};

\coordinate (zcbus3) at (-10.8,-2.7);

\draw[-,very thick] (GcEq.west) -- (zcbus3);

\draw[->,very thick] (zcbus3) |- (Leq.south);


\draw[->,very thick] (Ty.south) -- (Leq.north);


\node[physop] (GcRaw) at (-3.9,-5.3)
{
    $\*A_{c}$
};

\draw[->,very thick]
([xshift=-.75cm]sCT.south) -- node[right] {$\hat{\*x}_0$}
([xshift=.55cm]GcRaw.north);

\node[rawloss] (Lraw) at (-0.6,-5.3)
{
    $\displaystyle
    \mathcal{L}_{\mathrm{RawDC}} = \frac{1}{M_c}
    \left\| \*W_{c}^{1/2}(\mathbf{y}_{c} - \*A_c
    \hat{\mathbf{x}}_0) \right\|_2^2$
};

\draw[->,very thick] (GcRaw.east) -- (Lraw.west);

\draw[->,very thick] (yc.west) -- node[above] {measured} (Lraw.east);


\begin{pgfonlayer}{background}

\node[
    background,
    fit=(z0)(ztm1)(zT),
    fill=cyan!18,
    label=above:{
        \textbf{Forward diffusion in latent space}
    }
] {};

\node[
    background,
    fit=(epsT)(epsTm1)(eps1),
    fill=red!15,
    label=above:{
        \textbf{Conditional reverse diffusion}
    }
] {};

\node[
    rectangle,
    rounded corners=4mm,
    draw=blue!50!black,
    dashed,
    thick,
    fill=blue!4,
    inner sep=.160cm,
    fit=(GcRef)(Ty)(sCTrot)(GcEq)(Leq),
    label=below:{\textbf{\shortstack{CBCT geometry\\induced equivariance}}}
] {};

\node[
    rectangle,
    rounded corners=4mm,
    draw=green!50!black,
    dashed,
    thick,
    fill=green!4,
    inner sep=0.18cm,
    fit=(GcRaw)(Lraw)(yc),
    label=below:{
        \textbf{Noisy raw data consistency}
    }
] {};

\end{pgfonlayer}

\end{tikzpicture}
}

\caption{Overview of the proposed EPC-RawDC framework. The reference CT $\*x_0$ and input CBCT $\*x_c$ are encoded into volumetric latent representations, and conditional 3D diffusion produces the synthesized CT $\hat{\*x}_0$. Geometry equivariance links an image domain rotation $\*R_{\phi}$ to the corresponding angular translation $\*T_{\phi}$ in projection space, while RawDC constrains $\hat{\*x}_0$ using the measured CBCT projections. 
}\label{fig:EPC_RAWDC_workflow}

\end{figure*}

Building on the statistical measurement model and the scanner geometry induced equivariance established in Section~\ref{subsec:cbct_equivariance}, EPC-RawDC combines conditional 3D latent diffusion with two complementary projection domain physics terms. The first provides geometry transformed reference projection supervision derived from the rotational equivariance of the acquisition operator, whereas RawDC uses the noisy projection measurements of the individual scan as training supervision. 

Let $\*x_0$ denote the reference CT volume, $\*x_c$ the spatially aligned CBCT volume, and $\*A_c$ the scanner specific CBCT forward projector at the native detector resolution.

\subsection{Conditional 3D Latent Diffusion Model} \label{subsec:latent_diffusion}

Direct diffusion on full resolution 3D volumes is computationally demanding. We therefore perform the generative process in a compact volumetric latent space, following the latent diffusion principle \cite{rombach2022ldm} and the conditional formulation introduced in EPC-3D-Diff. A shared encoder $\mathcal{E}_{\psi}$ maps the paired CT and CBCT volumes to
\begin{equation} \label{eq:latent_encoding}
    \mathbf{z}_{0} = \mathcal{E}_{\psi}(\mathbf{x}_{0}), \qquad \mathbf{z}_{c} = \mathcal{E}_{\psi}(\mathbf{x}_{c}),
\end{equation}
with $\mathbf{z}_{0},\mathbf{z}_{c}\in\mathbb{R}^{C\times N'\times M'\times S},\, N'<N,\, M'<M$, and $C$ the number of latent channels. The in plane dimensions are compressed while volumetric depth is retained, allowing the diffusion model to exploit inter slice context at lower memory cost. After autoencoder pretraining, the encoder $\mathcal{E}_{\psi}$ and decoder $\mathcal{D}_{\omega}$ are fixed during diffusion training.

A variance preserving diffusion process \cite{ho2020ddpm} is applied to the target CT latent $\mathbf{z}_{0}$. Defining $\alpha_t=1-\beta_t$ and $\bar{\alpha}_t=\prod_{s=1}^{t}\alpha_s$, the forward process is $q(\mathbf{z}_{t}\mid\mathbf{z}_{0}) = \mathcal{N}\left( \sqrt{\bar{\alpha}_{t}}\mathbf{z}_{0}, (1-\bar{\alpha}_{t})\mathbf{I} \right)$, or equivalently,
\begin{equation} \label{eq:latent_noising}
    \mathbf{z}_{t} = \sqrt{\bar{\alpha}_{t}}\mathbf{z}_{0} + \sqrt{1-\bar{\alpha}_{t}}\,\bm{\epsilon},
    \qquad \bm{\epsilon}\sim\mathcal{N}(\mathbf{0},\mathbf{I}).
\end{equation}

The reverse process is parameterized by a conditional 3D UNet $\bm{\epsilon}_{\theta}$ that receives the noisy CT latent, the diffusion time step, and the CBCT latent condition, $\hat{\bm{\epsilon}} = \bm{\epsilon}_{\theta} \left(\mathbf{z}_{t},t,\mathbf{z}_{c}\right)$. The corresponding estimate of the clean CT latent is
\begin{equation} \label{eq:z0_prediction}
    \hat{\mathbf{z}}_{0} = \frac{\mathbf{z}_{t} - \sqrt{1-\bar{\alpha}_{t}}\,
        \bm{\epsilon}_{\theta} (\mathbf{z}_{t},t,\mathbf{z}_{c})}{\sqrt{\bar{\alpha}_{t}}},
\end{equation}
which is decoded to the predicted synthetic CT, $\hat{\mathbf{x}}_{0} = \mathcal{D}_{\omega}
\left(\hat{\mathbf{z}}_{0}\right)$. The conditional denoising objective is
\begin{equation} \label{eq:ddpm_loss_epc_rawdc}
    \mathcal{L}_{\mathrm{DDPM}} = \mathbb{E}_{t,\mathbf{z}_{0},\bm{\epsilon}} \left[
    \left\| \bm{\epsilon} - \bm{\epsilon}_{\theta} (\mathbf{z}_{t},t,\mathbf{z}_{c})
    \right\|_{2}^{2} \right].
\end{equation}

Importantly, the physics constraints introduced below are evaluated on $\hat{\mathbf{x}}_{0}$ rather than directly on the noisy latent $\mathbf{z}_{t}$. Consequently, the acquisition model acts on a physical
image domain estimate of the attenuation volume while the generative prior is learned efficiently in latent space.

\subsection{Geometry Derived Equivariance Projection Constraint} \label{subsec:eq_loss}

As established in Section~\ref{subsec:cbct_equivariance}, a rotation $\mathbf{R}_{\phi}$ of the attenuation volume about the CBCT scanner axis induces a corresponding angular translation $\mathbf{T}_{\phi}$ in the projection domain. This property is inherited from the circular source detector trajectory and is therefore a deterministic property of the acquisition operator, rather than an image augmentation heuristic. To distinguish the geometry transformed reference target from the noisy CBCT measurements used by RawDC, we define the noise free projection of the reference CT under the CBCT geometry as $\bar{\*y}_{0\rightarrow c} := \*A_c\*x_{0}$.  

For an admissible rotation $\phi_i$, the forward projection of the rotated prediction should agree with the corresponding angular translation of this reference projection 
\begin{equation} \label{eq:predicted_equivariance}
    \*A_{c} \*R_{\phi_i} \hat{\*x}_{0} \simeq \*T_{\phi_i}\bar{\*y}_{0\rightarrow c}.
\end{equation}

The geometry equivariant projection loss is therefore
\begin{equation}\label{eq:eq_loss_epc_rawdc}
\mathcal{L}_{\mathrm{eq}} =
\frac{1}{N_{\phi}} \sum_{i=1}^{N_{\phi}}
\left\|\mathbf{A}_{c}\mathbf{R}_{\phi_i}\hat{\mathbf{x}}_{0} - \mathbf{T}_{\phi_i}\bar{\mathbf{y}}_{0\rightarrow c} \right\|_{2}^{2},
\end{equation}
where $N_{\phi}$ is the number of sampled rotations. 
To make the role of Eq.~\eqref{eq:eq_loss_epc_rawdc} explicit, define the discrete operator residual
$\boldsymbol{\delta}_{\phi}(\mathbf{x})
=\mathbf{A}_{c}\mathbf{R}_{\phi}\mathbf{x}
-\mathbf{T}_{\phi}\mathbf{A}_{c}\mathbf{x}$.
Then
\begin{equation}\label{eq:eq_decomposition}
\mathbf{A}_{c}\mathbf{R}_{\phi}\hat{\mathbf{x}}_{0}
-\mathbf{T}_{\phi}\mathbf{A}_{c}\mathbf{x}_{0}
=
\mathbf{T}_{\phi}\mathbf{A}_{c}
(\hat{\mathbf{x}}_{0}-\mathbf{x}_{0})
+
\boldsymbol{\delta}_{\phi}(\hat{\mathbf{x}}_{0}).
\end{equation}
In the ideal continuous circular geometry, $\boldsymbol{\delta}_{\phi}=0$; if $\mathbf{T}_{\phi}$ is norm preserving, Eq.~\eqref{eq:eq_loss_epc_rawdc} reduces to reference projection supervision
$\|\mathbf{A}_{c}(\hat{\mathbf{x}}_{0}-\mathbf{x}_{0})\|_{2}^{2}$.
Accordingly, we interpret $\mathcal{L}_{\mathrm{eq}}$ as geometry transformed projection supervision derived from equivariance of the CBCT forward operator, rather than as a direct constraint on equivariance of the learned synthesis mapping. In the discrete implementation, finite angular and detector sampling together with spatial and angular interpolation give $\boldsymbol{\delta}_{\phi}\neq0$. Therefore, the transformed formulation is not exactly equivalent to unrotated projection matching. The notation $\mathcal{L}_{\mathrm{eq}}$ is retained for continuity with EPC-3D-Diff.

\subsection{Raw Data Consistency Constraint} \label{subsec:rawdc}

The equivariant term uses a transformed reference projection target. Therefore, the predicted sCT volume is not necessarily consistent with the raw noisy measurements observed for a particular patient. To introduce scan specific information, we additionally constrain $\hat{\*x}_0$ using the measured CBCT projection data $\*y_c$. Following the model in Section~\ref{subsec:measurement_model}, the underlying photon counts obey Poisson statistics, while the post-log measurements are approximated by $\mathbf{y}_{c} = \mathbf{A}_{c}\mathbf{x} + \mathbf{n}_{c}, \;\mathbf{n}_{c} \sim \mathcal{N}(\mathbf{0},\boldsymbol{\Sigma}_{c})$.
The corresponding general weighted least squares RawDC data fidelity term is
\begin{equation} \label{eq:rawdc_wls_general}
    \mathcal{L}_{\mathrm{RawDC}} = \frac{1}{M_c} \left\| \*W_{c}^{1/2} \left(\*y_{c} - \*A_{c}\hat{\*x}_{0} \right) \right\|_{2}^{2}, \; \*W_c = \boldsymbol{\Sigma}_{c}^{-1}
\end{equation}
where $M_c$ denotes the number of projection samples used in the loss. The normalization makes the magnitude of the loss less dependent on the number of views and detector elements.

RawDC therefore complements Eq.~\eqref{eq:eq_loss_epc_rawdc} by changing the projection domain target from the transformed noise free reference projection to the particular noisy realization $\*y_c$. Both terms use the same scanner specific forward model, but the geometry derived term provides reference based projection supervision whereas RawDC introduces patient and scan specific measurement evidence.

\subsection{Overall Training Objective} \label{subsec:total_objective}

In addition to the conditional diffusion objective, image domain regularization is retained from EPC-3D-Diff to preserve quantitative voxel accuracy and local anatomical structure. For the predicted clean
CT $\hat{\mathbf{x}}_{0}$, we use
\begin{align}
    \mathcal{L}_{1} &= \left\| \hat{\mathbf{x}}_{0} - \mathbf{x}_{0} \right\|_{1}, \label{eq:l1_loss_new}
    \\
    \mathcal{L}_{\mathrm{edge}} &= \left\| \nabla\hat{\mathbf{x}}_{0} - \nabla\mathbf{x}_{0} \right\|_{1},
    \quad \mathcal{L}_{\mathrm{lap}} = \left\| \Delta\hat{\mathbf{x}}_{0} - \Delta\mathbf{x}_{0} \right\|_{1}, \nonumber
\end{align}
where $\nabla$ and $\Delta$ denote the 3D finite difference gradient and Laplacian operators, respectively.

The complete EPC-RawDC training objective is
\begin{equation}\label{eq:epc_rawdc_total_loss}
\begin{aligned}
    \mathcal{L}_{\mathrm{total}} ={}& \mathcal{L}_{\mathrm{DDPM}} + \lambda_{1}\mathcal{L}_{1} +
    \lambda_{2}\mathcal{L}_{\mathrm{edge}} + \lambda_{3}\mathcal{L}_{\mathrm{lap}} \\
    &+ \lambda_{\mathrm{eq}}\mathcal{L}_{\mathrm{eq}} + \lambda_{\mathrm{RawDC}} \mathcal{L}_{\mathrm{RawDC}},
\end{aligned}
\end{equation}
where $\lambda_{1}$, $\lambda_{2}$, $\lambda_{3}$, $\lambda_{\mathrm{eq}}$, and $\lambda_{\mathrm{RawDC}}$ control the relative contribution of the corresponding constraints.

The terms in Eq.~\eqref{eq:epc_rawdc_total_loss} act at complementary levels. $\mathcal{L}_{\mathrm{DDPM}}$ learns the conditional latent CT distribution, the image domain losses preserve voxel and structural fidelity, $\mathcal{L}_{\mathrm{eq}}$ provides geometry transformed reference projection supervision derived from scanner equivariance, and $\mathcal{L}_{\mathrm{RawDC}}$ provides scan specific supervision from noisy projection measurements. Consequently, EPC-RawDC combines a learned 3D generative prior with scanner derived geometry and patient specific measurement information within a single training objective.  

During inference, only the reconstructed CBCT volume $\mathbf{x}_{c}$ is required as conditioning input. Sampling is performed in latent space and the final latent estimate is decoded to obtain $\hat{\mathbf{x}}_{0}$. The projection operators and raw measurements are used only as training supervision; therefore, projection agreement for a new patient is encouraged by the learned conditional model rather than explicitly enforced at inference.

\section{Experimental Data and Acquisition Geometry}
\label{sec:data}

The framework was evaluated on two independent head and neck cohorts SynthRAD2025 \cite{thummerer2025synthrad} and JUST with distinct clinical populations and CBCT geometries as in Table \ref{tab:datasets}. 
All volumes were resampled to a common grid of $50\times256\times256$ voxels with $1\times1\times3$ mm spacing. Image intensities were clipped to $[-1000,2000]$ HU and linearly mapped to $[-1,1]$. Training and testing were separated at the patient level.

\begin{table}[!h]
\centering
\caption{Summary of the clinical datasets and CBCT acquisition models used in the study.}
\label{tab:datasets}
\setlength{\tabcolsep}{5pt}
\renewcommand{\arraystretch}{1.15}
\begin{tabular}{lcc}
\hline
 & \textbf{SynthRAD2025} & \textbf{JUST} \\
\hline
Source & Multicenter clinical CT & Clinical CT and CBCT \\

Patients
& 96 train / 24 test & 32 train / 8 test \\

3D grid & $50\times256\times256$ & $50\times256\times256$ \\

Voxel spacing & $1\times1\times3$ mm & $1\times1\times3$ mm \\

CBCT geometry & Varian & Elekta \\

Detector & $768\times1024$ & $512\times512$ \\

Projection data & Poisson $I_0=10^{4}$ & Clinical raw projections \\
\hline
\end{tabular}
\end{table}

\begin{table*}[!t]
\centering
\caption{Average and Standard Deviation quantitative comparison at patient level: DDPM, EPC-3D-Diff, RawDC, EPC-RawDC, 3D ResUNet, 3D CycleGAN, on Single Domain SynthRAD2025 and JUST Test Sets for axial orientation.}
\label{tab:single_domain_results}
\begin{tabular}{c|l|c|ccc>{\columncolor{gray!30}}c|cc}
\hline
\rowcolor{green!20}
\cellcolor{white}\multirow{2}{*}{\textbf{Data}} & \cellcolor{white}\multirow{2}{*}{\textbf{Metrics}} & 

\cellcolor{red!20}\multirow{2}{*}{%
  \begin{tikzpicture}[overlay,remember picture]
    \fill[red!20] (-.21,.1) rectangle (6.85em,-0.89em);
  \end{tikzpicture} %
  \textbf{CBCT vs CT}}

& \multicolumn{4}{c|}{\textbf{CT vs Synth CT (Ablation)}} & \multicolumn{2}{c}{\textbf{CT vs Synth CT (Comparison)}}\\
\cline{4-9}
& &  & \textbf{DDPM} & \textbf{EPC-3D-Diff} & \textbf{RawDC} & \textbf{EPC-RawDC} &  \textbf{3D ResUNet} & \textbf{3D CycleGAN}  \\
\hline
\hline
\multirow{3}{1.5cm}{\textbf{SynthRAD25}}  & SSIM  & \cellcolor{red!20}$0.66 \pm 0.03$ & $0.84 \pm 0.04$ & $0.96 \pm 0.02$ & $0.96 \pm 0.02$  & $\mathbf{0.97} \pm \mathbf{0.02}$ &  $0.91 \pm 0.02$ & $0.95 \pm 0.02$ \\
& PSNR (dB) & \cellcolor{red!20}$27.65 \pm 0.96$ & $36.52 \pm 1.18$ & $37.94 \pm 1.64$ & $39.12 \pm 1.72$  & $\mathbf{39.18} \pm \mathbf{1.82}$ & $34.74 \pm 0.83$ &  $35.91 \pm 1.59$  \\
& MAE (HU) & \cellcolor{red!20}$37.53 \pm 3.50$ & $10.88 \pm 2.65$ & $8.71 \pm 3.20$ & $8.70 \pm 2.80$ & $\mathbf{8.67 \pm 2.81}$ & $13.81 \pm 1.59$ & $12.00 \pm 3.31$ \\
\hline
\hline
\multirow{3}{1.5cm}{\centering\textbf{JUST}}  & SSIM & \cellcolor{red!20}$0.60 \pm 0.02$ & $0.92 \pm 0.005$ & $0.96 \pm 0.01$ & $0.96 \pm 0.02$  & $\mathbf{0.97} \pm \mathbf{0.01}$ &  $0.95 \pm 0.01$   & $0.95 \pm 0.01$ \\
& PSNR (dB)       & \cellcolor{red!20}$25.50 \pm 0.54$ & $35.90 \pm 1.09$ & $37.90 \pm 3.65$ & $37.99 \pm 3.78$  & $\mathbf{38.00} \pm \mathbf{3.68}$ &  $35.59 \pm 1.03$ & $36.06 \pm 0.82$  \\
& MAE (HU) & \cellcolor{red!20}$49.26 \pm 3.73$ & $11.46 \pm 3.31$ & $9.83 \pm 2.80$ & $9.65 \pm 2.92$ & $\mathbf{9.28 \pm 2.91}$ & $11.32 \pm 1.68$ & $11.12 \pm 1.65$\\
\hline
\hline

\end{tabular}
\end{table*} 

\subsection{SynthRAD2025 Cohort}\label{subsec:synthrad_data}

SynthRAD2025 provides multicenter clinical CT and CBCT data for synthetic CT generation in radiotherapy \cite{thummerer2025synthrad}. Because raw projection measurements are not distributed with the dataset, the reference CT volumes were used to generate a controlled CBCT acquisition domain using a Varian cone beam geometry. The scanner forward operator was implemented with the ASTRA Toolbox \cite{van2016fast}. Projection measurements were generated from the CT attenuation volumes using Poisson photon statistics with an incident fluence of $I_0=10^{4}$ photons per ray, followed by FDK reconstruction. The native simulated detector consisted of $768\times1024$ elements. 

\subsection{JUST Clinical Cohort}\label{subsec:just_data}

The JUST cohort consists of paired clinical CT and CBCT examinations acquired at King Abdullah University Hospital, Jordan, with ethical approval Mar2026/191-38. In contrast to SynthRAD2025, the projection measurements used for the physics constraints were obtained from the clinical raw CBCT data. Therefore, the noise characteristics of the measured acquisition were retained. 
The corresponding Elekta acquisition model contains 220 projection views and a native $512\times512$ detector. 
The ASTRA Toolbox was used to implement the scanner specific forward operator employed by the geometry derived projection and RawDC constraints. The paired JUST CT and CBCT volumes were spatially aligned and resampled to the same grid to ensure voxelwise correspondence. Using a common image representation, while retaining the corresponding Varian and Elekta forward operators, allowed the same 3D diffusion architecture to be evaluated across both cohorts, without removing differences in scanner acquisition physics.

\section{Results}\label{sec:results}

\begin{figure*}[!h]
\centering
	\includegraphics[width=.95\textwidth]{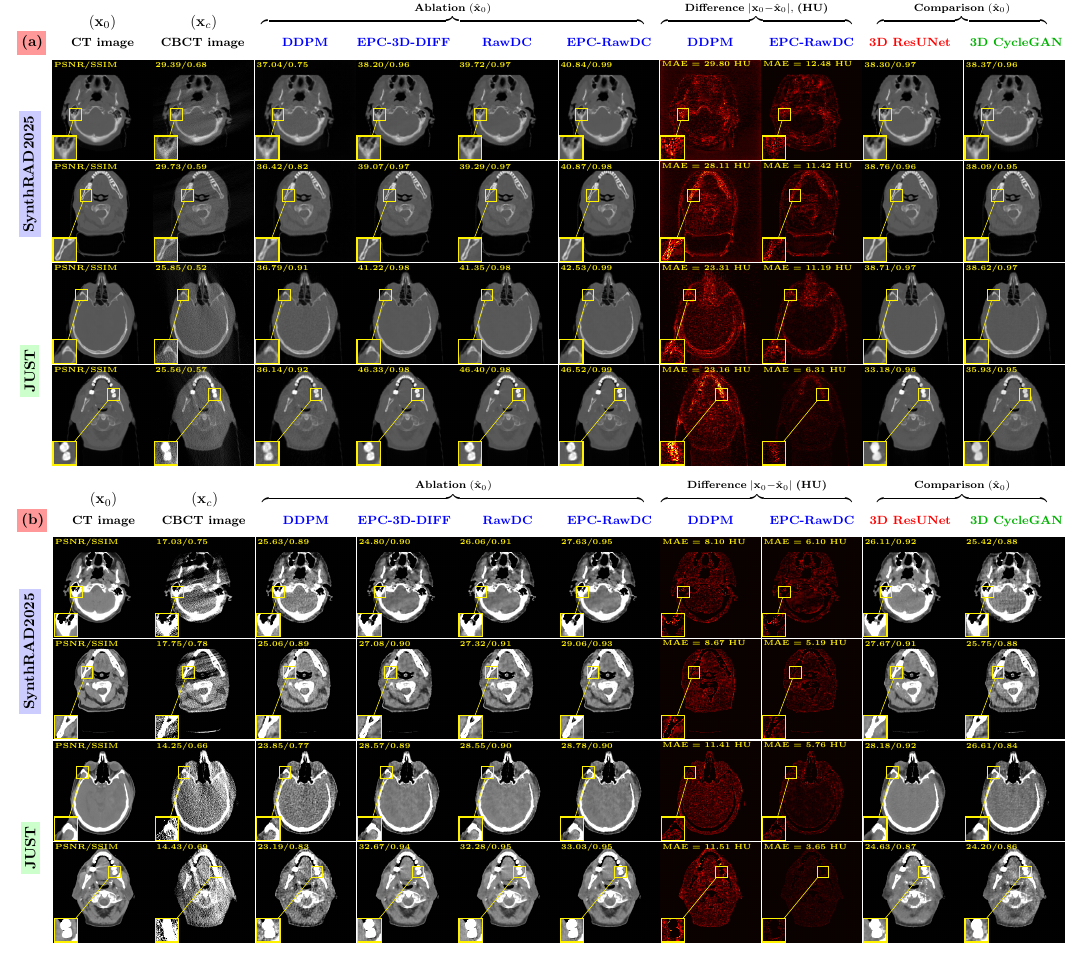}
	\caption{Qualitative comparison after single domain training for the SynthRAD2025 and JUST clinical test sets using (a) the full HU range and (b) the soft tissue window. Columns show the reference CT ($\*x_0$), CBCT ($\*x_c$), synthesized CT ($\hat{\*x}_0$) for the ablation and comparison methods, and the corresponding absolute error maps.
    }\label{fig:single_domain_visual}
\end{figure*}

We evaluated \method{} through four complementary experiments assessing: 1) component contributions under matched training and testing conditions; 2) generalization across independent clinical datasets and acquisition geometries; 3) joint multi domain training; and 4) volumetric 3D versus corresponding 2D modeling. Quantitative metrics were computed for each patient within the experiment. Images were clipped to $[-1000,2000]$ HU and normalized to $[0,1]$ for SSIM and PSNR, using a data range of $1$, while MAE was reported in HU. One value per metric was obtained for each patient, and cohort results are reported as mean $\pm$ standard deviation.

\subsection{Implementation Details}

The same network and optimization configuration was used for SynthRAD2025 and JUST, with the dataset and scanner specific geometry changed accordingly. Input volumes were encoded into 4 channel latent representations of size $4\times50\times128\times128$, preserving the complete through plane dimension. The pretrained encoder and decoder, implemented with 4 latent, 32 hidden channels and 500 epochs, were fixed during diffusion training. The conditional 3D UNet used 64 base channels and a 256 dimensional sinusoidal time embedding. Diffusion was performed over $1000$ time steps using a linear variance schedule from $\beta_1=10^{-4}$ to $\beta_T=5\times10^{-3}$. Training used Adam with a learning rate of $5\times10^{-6}$ and batch size one. The rotational projection constraint used one randomly signed rotation of $2^\circ$--$10^\circ$. Validation sampling used a 100 step DDIM procedure.

\subsection{Single Domain Performance and Component Ablation}
\label{subsec:results_single_domain}

We first evaluated all methods in the single domain setting, where training and testing were performed on the same clinical cohort using the same CBCT acquisition geometry. This experiment assesses reconstruction performance before introducing scanner or domain mismatch. 

For the JUST cohort, the models were trained using the Elekta acquisition geometry and evaluated on eight independent patients that were not included during training. For SynthRAD2025, training and testing were performed using the corresponding acquisition setting, with 24 independent patients included in the final evaluation. In both cohorts, the data split was defined at the patient level to prevent information leakage between training and testing. 

The ablation study included four diffusion based variants: DDPM  \cite{peng2024cbctddpm}, EPC-3D-Diff \cite{altalib2026epc3d}, RawDC and the complete EPC-RawDC model. The comparison models included a volumetric residual UNet (3D ResUNet), implemented as a 3D extension of residual UNet architectures previously investigated for CBCT based synthetic CT generation \cite{Caglar2025ResUNet}. 
A volumetric CycleGAN was also implemented using 3D convolutional generators and discriminators, consistent with recent applications in medical image synthesis \cite{MolinaMaza2025CBCT}.

Table~\ref{tab:single_domain_results} summarizes the patient level quantitative results for both cohorts. All synthesis methods substantially improved the input CBCT, with the physics informed diffusion variants showing the strongest overall performance. EPC-3D-Diff and RawDC both improved upon DDPM, indicating that geometry derived reference projection supervision and scan specific projection supervision each contribute to reconstruction quality. Their combination in EPC-RawDC provided the best overall results in both datasets, reaching an SSIM of $0.97$ in both cohorts and the highest PSNR of $39.18$ dB for SynthRAD2025 and $38.00$ dB for JUST. Compared with DDPM, EPC-RawDC reduced the mean MAE by approximately $20.3\%$ for SynthRAD2025 and $19.0\%$ for JUST. 

The comparison models also improved substantially over the input CBCT, but their overall performance remained below EPC-RawDC. Across both datasets, EPC-RawDC achieved higher PSNR and lower MAE than 3D UNet and 3D CycleGAN, while also providing the highest SSIM. The consistent behaviour across SynthRAD2025 and JUST suggests that the benefit of combining equivariant projection supervision with RawDC is maintained across both acquisition settings. 

These findings support the complementary use of the two projection domain terms. EPC-3D-Diff uses geometry transformed supervision from the reference CT, whereas RawDC uses the projection measurements of the acquisition. Their combination produced the best mean performance, although the patient level analysis below shows that some incremental differences relative to RawDC were not statistically significant.  

Figure~\ref{fig:single_domain_visual} presents representative axial examples from the JUST and SynthRAD2025 test cohorts. For each case, the reference CT, input CBCT, synthesized CT images, and absolute error maps are shown. The examples illustrate reconstruction behaviour in soft tissue, bone, air cavities, and regions affected by CBCT intensity artifacts. Images are displayed using both the full HU range $[-1000, 2000]$ HU and the soft tissue range $[-160, 240]$ HU. 
The full HU range provides an overall view of the reconstructed anatomy, whereas the soft tissue window enhances smaller intensity differences. MAE values computed within these windows are therefore not directly comparable, as the full HU range spans substantially larger intensity variations.

\subsection{Cross Dataset Generalization under Fixed Geometry
} \label{subsec:results_generalization}

Cross dataset generalization was evaluated in both transfer directions while preserving the acquisition geometry used during training. The model trained on SynthRAD2025 with the Varian geometry was evaluated on 8 subjects from JUST dataset represented using the same Varian forward model, whereas the model trained on JUST with the Elekta geometry was evaluated on 24 SynthRAD2025 patients represented using the corresponding Elekta forward model. The results are reported in Tables~\ref{tab:synth_to_just_generalization} and \ref{tab:just_to_synth_generalization}.

For transfer from SynthRAD2025 to JUST, EPC-RawDC showed a clear improvement over DDPM, increasing SSIM by approximately $5.7\%$, increasing PSNR by $1.32$ dB, and reducing MAE by $23.7\%$. Relative to RawDC, EPC-RawDC maintained the same SSIM at the reported precision, while improving PSNR by $0.47$ dB and reducing MAE by approximately $5.6\%$. This indicates that the combined constraints remained beneficial when the model was applied to an independent clinical population.

The reverse transfer from JUST to SynthRAD2025 showed the same overall trend, although the improvement was smaller. Compared with DDPM, EPC-RawDC increased SSIM by approximately $5.6\%$ and reduced MAE by $6.8\%$, while PSNR increased by $0.04$ dB. Compared with RawDC, EPC-RawDC further improved SSIM by approximately $1.1\%$ and reduced MAE by $2.5\%$. The consistent performance in both transfer directions suggests that the physics constrained framework improves robustness to changes in patient population while the CBCT forward geometry is maintained.

\begin{table}[!h]
\centering
\caption{Cross dataset generalization from SynthRAD2025 training to the JUST test cohort using Varian geometry.
}\label{tab:synth_to_just_generalization}

\setlength{\tabcolsep}{4.5pt}
\renewcommand{\arraystretch}{1.15}

\begin{tabular}{lcccc}
\hline
\rowcolor{green!20}  \textbf{Method} &
\textbf{SSIM} $\uparrow$ &
\textbf{PSNR (dB)} $\uparrow$ &
\textbf{MAE (HU)} $\downarrow$ \\
\hline

\rowcolor{red!20} \textbf{CBCT vs CT}
& $0.63 \pm 0.02$
& $27.88 \pm 0.66$
& $38.84 \pm 3.11$ \\

DDPM
& $0.87 \pm 0.01$
& $34.80 \pm 0.56$
& $16.70 \pm 0.88$ \\

RawDC
& $0.92 \pm 0.02$
& $35.65 \pm 0.64$
& $13.51 \pm 1.27$ \\

\rowcolor{gray!30}\textbf{EPC-RawDC}
& $\mathbf{0.92 \pm 0.02}$
& $\mathbf{36.12 \pm 0.63}$
& $\mathbf{12.75 \pm 1.31}$ \\

\hline
\end{tabular}
\end{table}

\begin{table}[!h]
\centering
\caption{Cross dataset generalization from JUST training to SynthRAD2025 test cohort using Elekta geometry.
}
\label{tab:just_to_synth_generalization}

\setlength{\tabcolsep}{5pt}
\renewcommand{\arraystretch}{1.15}

\begin{tabular}{lccc}
\hline
\rowcolor{green!20} \textbf{Method} &
\textbf{SSIM} $\uparrow$ &
\textbf{PSNR (dB)} $\uparrow$ &
\textbf{MAE (HU)} $\downarrow$ \\
\hline

\rowcolor{red!20} \textbf{CBCT vs CT}
& $0.57 \pm 0.06$
& $25.74 \pm 0.82$
& $47.89 \pm 7.16$ \\

DDPM
& $0.90 \pm 0.01$
& $33.98 \pm 1.39$
& $16.23 \pm 3.64$ \\

RawDC
& $0.94 \pm 0.01$
& $33.98 \pm 1.48$
& $15.51 \pm 3.91$ \\

\rowcolor{gray!30} \textbf{EPC-RawDC}
& $\mathbf{0.95 \pm 0.01}$
& $\mathbf{34.02 \pm 1.51}$
& $\mathbf{15.13 \pm 4.04}$ \\

\hline
\end{tabular}
\end{table}


\subsection{Joint Multi Domain Training} \label{subsec:results_combined_domain}

To assess whether a single model could accommodate heterogeneous acquisition physics, EPC-RawDC was trained jointly using the SynthRAD2025 and JUST cohorts, with each sample retaining its  corresponding scanner specific forward operator.
The shared model was then evaluated independently on the held out test cohort from each domain.

As shown in Table~\ref{tab:joint_domain_results}, EPC-RawDC consistently improved upon DDPM after joint training. On SynthRAD2025, EPC-RawDC increased SSIM by approximately $1.1\%$, improved PSNR by $0.49$ dB, and reduced MAE by $6.5\%$. A similar trend was observed on JUST, where SSIM increased by approximately $1.1\%$, PSNR improved by $0.35$ dB, and MAE decreased by $5.8\%$.

\begin{table}[!h]
\centering
\caption{Performance of the jointly trained models on the held out SynthRAD2025 and JUST test cohorts. The models were trained using both datasets and evaluated on each domain.
}\label{tab:joint_domain_results}

\setlength{\tabcolsep}{4pt}
\renewcommand{\arraystretch}{1.27}

\begin{tabular}{clccc}
\hline
\rowcolor{green!20} &
\textbf{Method} &
\textbf{SSIM} $\uparrow$ &
\textbf{PSNR (dB)} $\uparrow$ &
\textbf{MAE (HU)} $\downarrow$ \\
\hline

\multirow[c]{3}{*}{%
\rotatebox[origin=c]{90}{%
\colorbox{blue!15}{\scriptsize\bfseries RAD2025}%
}}
& \cellcolor{red!20}CBCT vs CT
& \cellcolor{red!20}$0.66 \pm 0.03$
& \cellcolor{red!20}$27.65 \pm 0.96$
& \cellcolor{red!20}$37.53 \pm 3.50$ \\

& DDPM
& $0.94 \pm 0.02$
& $34.08 \pm 1.03$
& $16.86 \pm 3.36$ \\

& \cellcolor{gray!30}\textbf{EPC-RawDC}
& \cellcolor{gray!30}$\mathbf{0.95 \pm 0.02}$
& \cellcolor{gray!30}$\mathbf{34.57 \pm 1.06}$
& \cellcolor{gray!30}$\mathbf{15.76 \pm 3.16}$ \\
\hline

\multirow[c]{3}{*}{%
\rotatebox[origin=c]{90}{%
\colorbox{green!15}{\scriptsize\bfseries JUST}%
}}
& \cellcolor{red!20}CBCT vs CT
& \cellcolor{red!20}$0.60 \pm 0.02$
& \cellcolor{red!20}$25.50 \pm 0.54$
& \cellcolor{red!20}$49.26 \pm 3.73$ \\

& DDPM
& $0.94 \pm 0.01$
& $34.19 \pm 1.19$
& $16.65 \pm 2.60$ \\

& \cellcolor{gray!30}\textbf{EPC-RawDC}
& \cellcolor{gray!30}$\mathbf{0.95 \pm 0.01}$
& \cellcolor{gray!30}$\mathbf{34.54 \pm 1.41}$
& \cellcolor{gray!30}$\mathbf{15.68 \pm 2.76}$ \\
\hline

\end{tabular}
\end{table}

The comparison with domain specific training in Table~\ref{tab:combined_domain_results} showed a clear advantage for specialization to a single acquisition domain. Relative to the corresponding single domain EPC-RawDC models, joint training reduced SSIM by approximately $2.1\%$ in both cohorts and decreased PSNR by $4.61$ dB on SynthRAD2025 and $3.46$ dB on JUST. MAE increased by approximately $81.8\%$ on SynthRAD2025 and $69.0\%$ on JUST. These results indicate that although a single EPC-RawDC model can accommodate heterogeneous scanner geometries, domain specific training provides greater structural similarity, contrast fidelity, and voxelwise intensity accuracy when sufficient scanner specific training data are available.

\begin{figure*}[!h]
	\includegraphics[width=\textwidth]{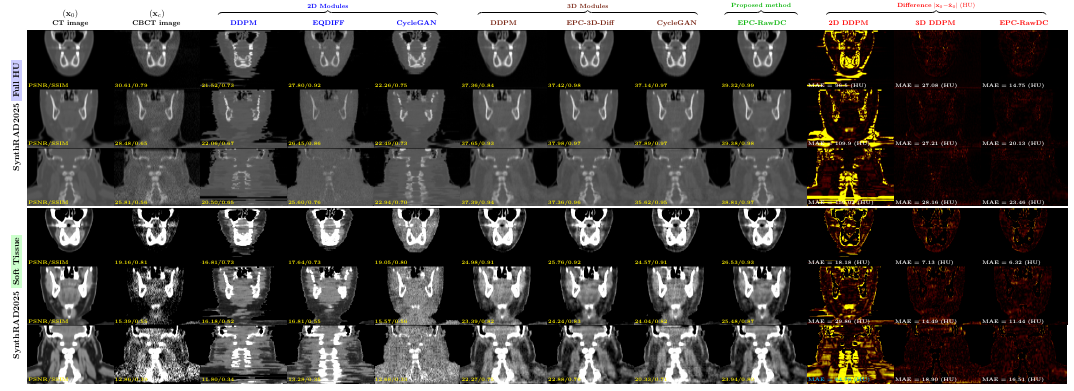}
	\caption{Coronal comparison of 2D and 3D CBCT to CT synthesis on representative SynthRAD2025 slices using full HU and soft tissue window. Columns show the reference CT, CBCT, 2D DDPM, 2D EqDiff, 2D CycleGAN, 3D DDPM, EPC-3D-Diff, 3D CycleGAN, and EPC-RawDC, and the absolute HU error for 2D DDPM, 3D DDPM, and EPC-RawDC. PSNR and SSIM are reported for the synthesis images and MAE for the error maps. 
    }
    \label{fig:2D_3D_domain_visual}
\end{figure*}

\begin{table}[!h]
\centering
\caption{Comparison of single domain and joint domain EPC-RawDC training across the SynthRAD2025 and JUST geometries.
}\label{tab:combined_domain_results}

\setlength{\tabcolsep}{4pt}
\renewcommand{\arraystretch}{1.26}

\begin{tabular}{llccc}
\hline
\rowcolor{green!20} \bfseries Test & \bfseries Training &
\bfseries SSIM $\uparrow$ & \bfseries PSNR $\uparrow$ & \bfseries MAE $\downarrow$ \\
\hline
\multirow[c]{2}{*}{\rotatebox[origin=c]{90}{\colorbox{blue!15}{\scriptsize\bfseries RAD}}}
& JUST + RAD2025  & $0.95 \pm 0.02$ & $34.57 \pm 1.06$ & $15.76 \pm 3.16$ \\
& RAD2025 only    & $0.97 \pm 0.02$ & $39.18 \pm 1.82$ & $8.67 \pm 2.81$ \\
\hline
\multirow[c]{2}{*}{\rotatebox[origin=c]{90}{\colorbox{green!15}{\scriptsize\bfseries JUST}}}
& JUST + RAD2025  & $0.95 \pm 0.01$ & $34.54 \pm 1.41$ & $15.68 \pm 2.76$ \\
& JUST only       & $0.97 \pm 0.01$ & $38.00 \pm 3.68$ & $9.28 \pm 2.91$ \\
\hline
\end{tabular}
\end{table}

\subsection{Volumetric 3D Versus Slice Wise 2D Modeling} \label{subsec}

The contribution of volumetric context was evaluated by comparing the proposed 3D framework with corresponding slice wise 2D synthesis methods. The 2D EqDiff model employs rotational equivariance directly in the image domain through a $C_4$ group equivariant denoising network \cite{altalib2026eqdiff}, whereas EPC-3D-Diff uses translated projection supervision derived from image rotational equivariance of the CBCT forward operator within a volumetric latent diffusion framework \cite{altalib2026epc3d}. EPC-RawDC further incorporates patient specific projection consistency into the 3D formulation. 

As summarized in Table~\ref{tab:2d3d_results}, the coronal 2D versus 3D evaluation across all 24 SynthRAD2025 test subjects showed a clear advantage for volumetric 3D modeling. 

Relative to 2D EqDiff, EPC-RawDC improved PSNR by $14.24$ dB and reduced MAE by approximately $75.1\%$. Within the 3D physics constrained models, EPC-RawDC further improved PSNR by $1.32$ dB and reduced MAE by $12.5\%$ relative to EPC-3D-Diff, while reducing MAE by $32.9\%$ relative to 3D DDPM. 

The coronal examples in Fig.~\ref{fig:2D_3D_domain_visual} further illustrate the effect of volumetric modeling on through plane consistency. The 2D approaches show stronger slice dependent intensity variation and discontinuity, particularly around the mandible, cervical spine, and surrounding soft tissue, whereas the 3D methods preserve more coherent anatomical structure across the coronal plane. EPC-RawDC provides the closest visual agreement with the reference CT in both the full HU and soft tissue displays. 
Across the displayed examples, EPC-RawDC achieved the highest PSNR and SSIM. Its error maps also showed the lowest MAE in every example compared with 2D and 3D DDPMs, with reduced residual errors around high contrast bone boundaries and soft tissue regions.

\begin{table}[!h]
\centering
\caption{Patient level comparison of 2D and 3D CBCT to CT synthesis across the SynthRAD2025 test subjects under the coronal evaluation protocol.}
\label{tab:2d3d_results}
\begin{tabular}{lccc}
\hline
\rowcolor{green!20}  \textbf{Method} &
\textbf{SSIM} $\uparrow$ &
\textbf{PSNR (dB)} $\uparrow$ &
\textbf{MAE (HU)} $\downarrow$ \\
\hline
\rowcolor{red!20} \textbf{CBCT vs CT}
& $0.66 \pm 0.04$ & $26.91 \pm 0.91$ & $43.02 \pm 6.50$ \\

2D DDPM & $0.66 \pm 0.12$ & $21.81 \pm 12.69$ & $58.35 \pm 26.91$ \\

2D EqDiff & $0.73 \pm 0.05$ & $24.28 \pm 1.03$ & $41.27 \pm 7.57$ \\

2D CycleGAN & $0.78 \pm 0.05$ & $26.55 \pm 1.50$ & $35.44 \pm 9.50$ \\
\hline 

3D DDPM & $0.89 \pm 0.02$ & $36.37 \pm 1.19$ & $15.28 \pm 1.92$ \\

3D CycleGAN & $0.94 \pm 0.02$ & $35.14 \pm 1.37$ & $14.27 \pm 3.45$ \\

3D EPC-3D-Diff & $0.95 \pm 0.02$ & $ 37.20 \pm 1.37$ & $11.72 \pm 3.10$ \\

\rowcolor{gray!30} \textbf{3D EPC-RawDC}
& $\mathbf{0.97 \pm 0.02}$ & $\mathbf{38.52 \pm 1.55}$ & $\mathbf{10.26 \pm 2.75}$ \\
\hline

\end{tabular}
\end{table}

These findings indicate that the principal improvement arises from exploiting volumetric anatomical context, while RawDC provides a further gain beyond the geometry derived projection term alone. Such volumetric consistency is particularly relevant to radiotherapy applications requiring spatially coherent image volumes for registration, contour propagation, and dose related analysis.

\begin{figure*}[!h]
\centering
	\includegraphics[width=.97\textwidth]{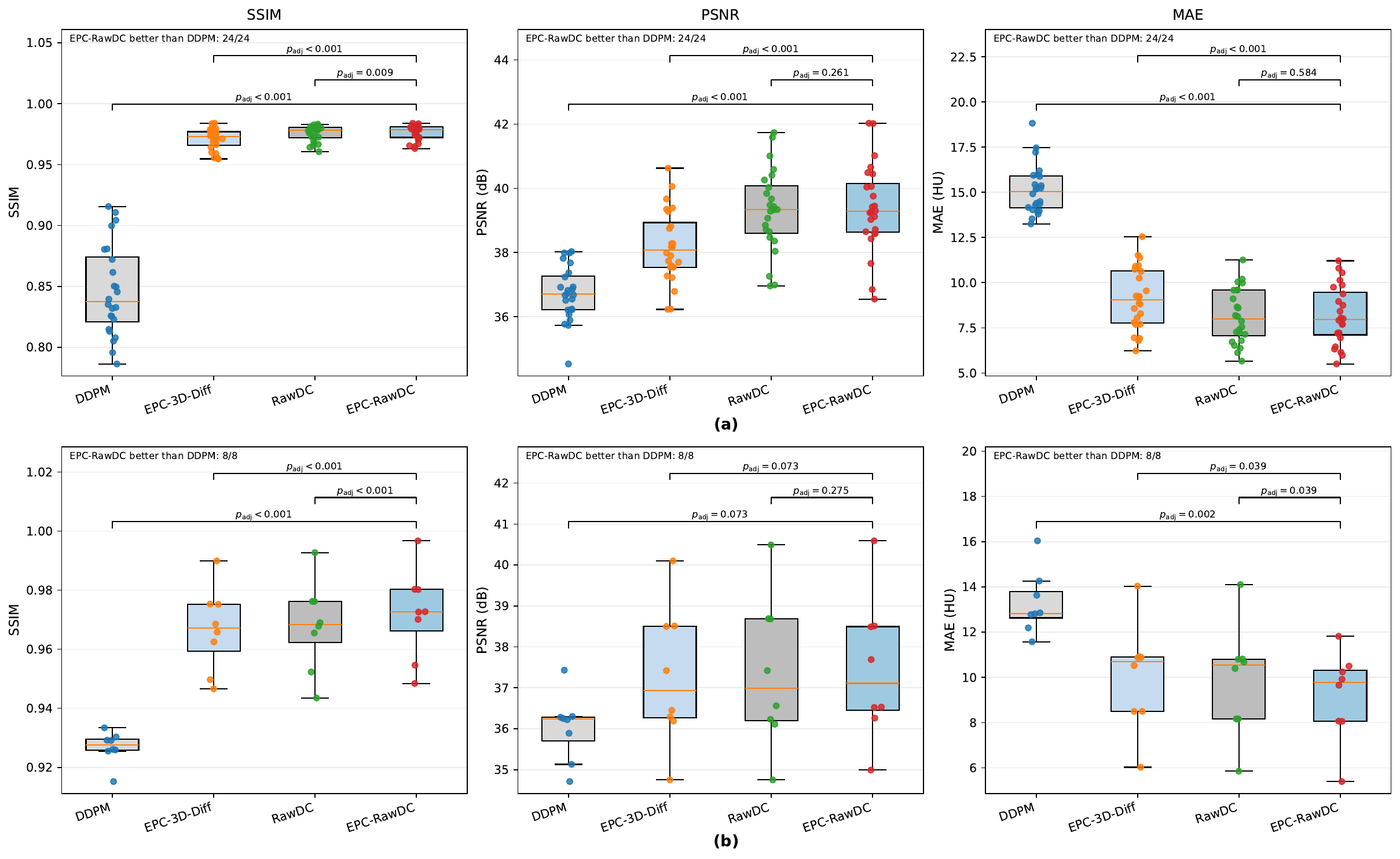}
	\caption{Patient level distributions of SSIM, PSNR, and MAE for (a) SynthRAD2025 and (b) JUST. DDPM, EPC-3D-Diff, RawDC, and EPC-RawDC are compared using matched patient level measurements. Boxes indicate the median and interquartile range, whiskers extend to 1.5 times the interquartile range, and individual points denote test patients. The annotation ``EPC-RawDC vs DDPM'' indicates the number of patients for which EPC-RawDC outperformed DDPM for the corresponding metric.
}
\label{fig:patient_boxplots}
\end{figure*}

\subsection{Physics Specific Consistency Analysis}
\label{subsec:results_physics}

Since the main contribution of EPC-RawDC is the incorporation of acquisition physics into training, we evaluated both projection domain physics terms directly in addition to conventional image quality metrics. 

Projection consistency was assessed using the same scanner specific forward model employed by RawDC during training. For each generated CT volume, the image was converted to attenuation coefficients and forward projected at the native detector resolution using $\mathbf{A}_{c}$. The projection consistency error was defined as $
\mathcal{E}_{\mathrm{proj}} = \frac{1}{M_c}\left\| \mathbf{y}_{c} - \mathbf{A}_{c} \hat{\mathbf{x}}_{0} \right\|_{2}^{2}$, where $M_c$ denotes the number of projection samples and lower values indicate greater agreement with the projection data.

As shown in Table~\ref{tab:physics_metrics}, DDPM produced a mean projection consistency error of $2.27\times10^{-3}$ across the 24 SynthRAD2025 test cases. This decreased to $1.05\times10^{-3}$ for EPC-3D-Diff and $1.02\times10^{-3}$ for EPC-RawDC, corresponding to a $55.0\%$ reduction relative to DDPM. The 3D CycleGAN achieved error of $1.11\times10^{-3}$. These results indicate that physics constrained models generate CT estimates that are more consistent with projection data than the unconstrained diffusion baseline.

\begin{table}[!h]
\centering
\caption{Direct evaluation of the physical consistency constraints across the 24 SynthRAD2025 test cases.}\label{tab:physics_metrics}
\begin{tabular}{lcc}
\hline
\rowcolor{green!20}\multirow{2}{*}{%
  \begin{tikzpicture}[overlay,remember picture]
    \fill[green!20] (-.21,.1) rectangle (7em,-1em);
  \end{tikzpicture}%
  \textbf{Method}} & \bfseries Projection error & \bfseries Equivariance error \\ 

&  \cellcolor{green!20}$\mathbf{(10^{-3})\;\downarrow}$ & \cellcolor{green!20}$\mathbf{(10^{-2})\;\downarrow}$ \\ 
\hline

DDPM
& $2.27 \pm 0.34$ & $7.18 \pm 1.69$ \\

3D CycleGAN & $1.11 \pm 0.72$ & $7.38 \pm 1.59$ \\

EPC-3D-Diff
& $1.05 \pm 0.71$ & $5.19 \pm 1.70$ \\

EPC-RawDC
& $\mathbf{1.02 \pm 0.50}$ & $\mathbf{4.21 \pm 1.67}$ \\

\hline
\end{tabular}
\end{table}

The geometry derived term was evaluated using the transformed projection residual. For each generated CT, the volume was rotated by a prescribed angle $\phi$ and forward projected using the same scanner specific operator $\mathbf{A}_{c}$. The result was compared with the corresponding angular translation of the reference CT projection:
$\mathcal{E}_{\mathrm{geo}} = \frac{1}{N_{\phi}}\sum_{i=1}^{N_{\phi}}
\left\|\mathbf{A}_{c}\mathbf{R}_{\phi_i}\hat{\mathbf{x}}_{0}
-\mathbf{T}_{\phi_i}\bar{\mathbf{y}}_{0\rightarrow c}\right\|_{2}^{2}$,
where $\bar{\mathbf{y}}_{0\rightarrow c}=\mathbf{A}_{c}\mathbf{x}_{0}$.
This metric quantifies agreement with the geometry transformed reference projection and is not a direct measurement of equivariance of the learned synthesis mapping. DDPM produced a mean geometry transformed projection error of $7.18\times10^{-2}$. EPC-3D-Diff reduced this value to $5.19\times10^{-2}$, corresponding to a $27.7\%$ reduction, while EPC-RawDC further reduced it to $4.21\times10^{-2}$, corresponding to reductions of $41.4\%$ relative to DDPM and $18.9\%$ relative to EPC-3D-Diff. The 3D CycleGAN produced $7.38\times10^{-2}$. Together, the two metrics assess complementary forms of projection domain consistency.

\subsection{Patient Level Statistical Analysis}
\label{subsec:results_statistics}

Paired patient level comparisons used matched measurements from each test subject. Normality of paired differences was assessed using Shapiro Wilk test, followed by a two sided paired $t$ test for normal differences or a two sided Wilcoxon signed rank test otherwise. Within each cohort, Holm Bonferroni correction was applied across 9 prespecified comparisons: EPC-RawDC vs DDPM, RawDC, and EPC-3D-Diff for SSIM, PSNR and MAE. Significance was defined as $p_{\mathrm{adj}}<0.05$. 

As shown in Fig.~\ref{fig:patient_boxplots}, EPC-RawDC consistently improved upon DDPM across both cohorts. On SynthRAD2025, EPC-RawDC outperformed DDPM in all 24 patients for SSIM, PSNR, and MAE, with all comparisons remaining significant after correction ($p_{\mathrm{adj}}<0.001$). EPC-RawDC also showed significant improvements over EPC-3D-Diff for all three metrics ($p_{\mathrm{adj}}<0.001$). Relative to RawDC, the improvement remained significant for SSIM ($p_{\mathrm{adj}}=0.009$), whereas the differences in PSNR and MAE were not significant ($p_{\mathrm{adj}}=0.261$ and $p_{\mathrm{adj}}=0.584$, respectively).

The JUST cohort showed a similar but more conservative pattern. EPC-RawDC outperformed DDPM in all 8 patients for all three metrics; however, after correction, the improvement was significant for SSIM ($p_{\mathrm{adj}}<0.001$) and MAE ($p_{\mathrm{adj}}=0.002$), but not for PSNR ($p_{\mathrm{adj}}=0.073$). Compared with the physics constrained alternatives, EPC-RawDC achieved significantly higher SSIM than both EPC-3D-Diff and RawDC (both $p_{\mathrm{adj}}<0.001$) and significantly lower MAE than both methods (both $p_{\mathrm{adj}}=0.039$), while the corresponding PSNR differences were not significant ($p_{\mathrm{adj}}=0.073$ versus EPC-3D-Diff and $p_{\mathrm{adj}}=0.275$ versus RawDC).

Overall, the patient level analysis confirms that the improvement over the unconstrained DDPM baseline is highly consistent across subjects. The smaller differences relative to EPC-3D-Diff and RawDC indicate that the incremental benefit of combining equivariant projection supervision with raw data consistency is most evident in structural similarity and, for the JUST cohort, in absolute intensity accuracy.  

\section{Conclusion}
\label{sec:conclusion}

This work introduced EPC-RawDC, a conditional 3D latent diffusion framework that integrates equivariance derived projection supervision with raw data consistency through a common acquisition CBCT forward operator. Across SynthRAD2025 and JUST, EPC-RawDC provided the best mean performance among the evaluated diffusion and comparison methods, reducing MAE relative to DDPM by 20.3\% and 19.0\%, respectively. Direct physics evaluation further demonstrated improved agreement with the projection measurements and geometry transformed reference projections. 
Cross dataset, joint domain, and volumetric experiments further characterized the generalization and multi domain behaviour of the framework, while domain specific training remained advantageous for quantitative accuracy. 
These findings support measurement grounded diffusion as a promising approach for quantitative CBCT to CT synthesis. Future work will investigate broader multicenter validation and the impact of the synthesized CT on downstream radiotherapy tasks, including dose calculation and adaptive treatment workflows.

\section*{Acknowledgement}

The authors thank Prof. Haytham Alewaidat, Prof. Khaled Alawneh, Eng. Ahmad Qandeel for help to collect the clinical JUST data and Mr. Mohammad Alhyari for the manuscript review.

\bibliographystyle{IEEEtran}
\bibliography{biblio}

\end{document}